\documentclass[conference]{IEEEtran}
\IEEEoverridecommandlockouts

\def\namerm{OpsMem}
\def\name{\textit{\namerm}\xspace}
\def\BibTeX{{\rm B\kern-.05em{\sc i\kern-.025em b}\kern-.08em
    T\kern-.1667em\lower.7ex\hbox{E}\kern-.125emX}}

\usepackage{amsmath}
\usepackage{amssymb}
\usepackage{xspace}
\usepackage{amsthm}

\usepackage{booktabs}
\usepackage{cite}
\usepackage[hidelinks]{hyperref}
\usepackage{multirow}
\usepackage{tabularx}
\usepackage{array}

\usepackage{threeparttable}
\usepackage{tablefootnote}

\usepackage{algorithm}
\usepackage{algorithmic}

\ifCLASSOPTIONcompsoc
  \usepackage[caption=false,font=normalsize,labelfont=sf,textfont=sf]{subfig}
\else
  \usepackage[caption=false,font=footnotesize]{subfig}
\fi

\usepackage{tikz}
\usetikzlibrary{decorations.pathreplacing,fit,intersections}

\theoremstyle{definition}

\begin{document}

\title{OpsMem: Dual-Memory Reasoning with \\ Cross-Memory Resonance for Failure Diagnosis}

\author{
    \IEEEauthorblockN{
    Yongqian Sun\IEEEauthorrefmark{2},
    Rongchen Gao\IEEEauthorrefmark{2},
    Yu Luo\IEEEauthorrefmark{2},
    Wenwei Gu\IEEEauthorrefmark{2},
    Shenglin Zhang\IEEEauthorrefmark{2}\IEEEauthorrefmark{1}\thanks{\IEEEauthorrefmark{1}Shenglin Zhang is the corresponding author.} \\
    Qingyi Guo\IEEEauthorrefmark{4},
    Qiuai Fu\IEEEauthorrefmark{3},
    Yaoliang Wu\IEEEauthorrefmark{3},
    Dan Pei\IEEEauthorrefmark{4}
    }
        
    \IEEEauthorblockA{
        \IEEEauthorrefmark{2}\textit{Nankai University} \quad
        \IEEEauthorrefmark{4}\textit{Tsinghua University} \quad 
        \IEEEauthorrefmark{3}\textit{Huawei Technologies Co., Ltd.}
    }
}

\maketitle

\setcounter{page}{1}
\thispagestyle{plain}
\pagestyle{plain}

\begin{abstract}
Failure diagnosis in modern software systems requires iterative evidence acquisition and hypothesis reasoning guided by operational experience. 
Existing LLM-based methods improve diagnosis through agentic reasoning or knowledge augmentation, but they often lack a mechanism to coordinate the evolving diagnostic state with operational experience during iterative diagnosis.
We propose \name, a dual-memory framework that maintains a short-term memory for the current diagnostic state and a long-term memory for reusable operational experience. 
\name uses cross-memory resonance to activate state-relevant long-term memory, conditions multi-agent diagnosis on the short-term and activated long-term memories, and consolidates reusable experience from solved incidents back into long-term memory. 
Experiments on a real-world Huawei microservice failure diagnosis dataset show that \name outperforms representative agentic-reasoning and knowledge-augmented baselines, improving Match and Relevant by up to 46.88\% and 18.39\% over the strongest baseline, respectively.

\end{abstract}



\section{Introduction}\label{sec:introduction}

\begin{figure*}[t]
    \centering
    \includegraphics[width=1\textwidth]{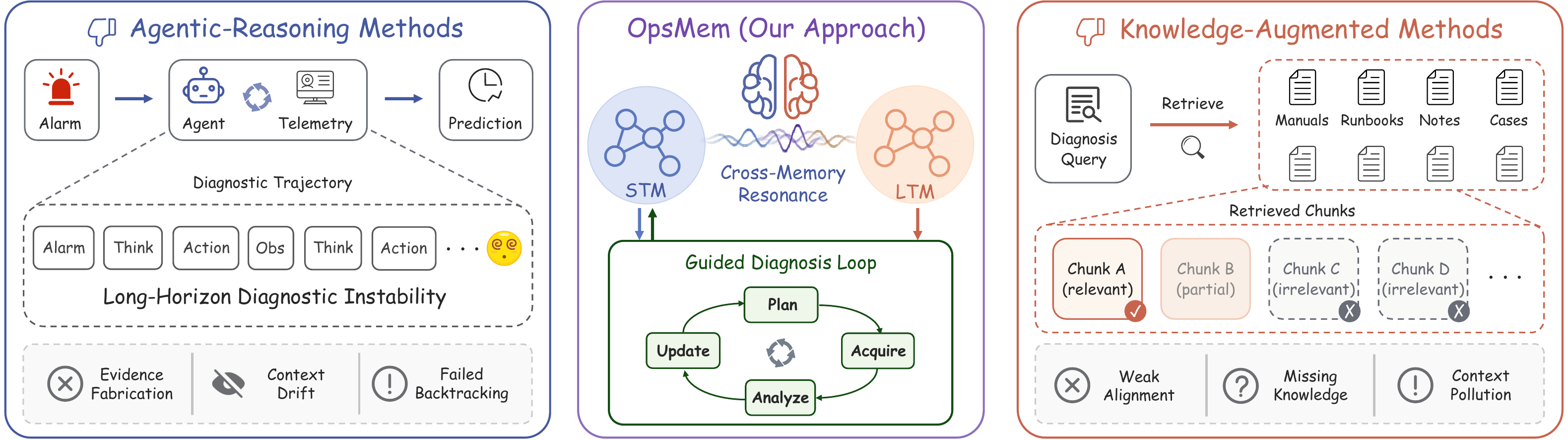}
    \vspace{-15pt}
    \caption{Motivation of OpsMem from a state–experience coupling perspective. Existing agentic-reasoning methods maintain diagnostic trajectories but suffer from long-horizon instability, whereas knowledge-augmented methods retrieve operational experience without continuously aligning it with the evolving diagnostic state. OpsMem couples short-term state memory and long-term experience memory via cross-memory resonance to guide diagnosis.}
    \label{fig:teaser}
    \vspace{-12pt}
\end{figure*}

Failure diagnosis is critical for maintaining the reliability of modern software systems, as engineers must quickly identify root causes and restore affected services when failures occur \cite{art}.
Traditionally, this task has relied heavily on manual inspection and expert experience, which is costly, time-consuming, and difficult to scale in today’s large-scale distributed environments \cite{OScope}. 
To automate this process, prior studies have explored data-driven methods based on machine learning and deep learning \cite{ MEPFL, diagfusion}. 
Although these methods have shown promise in specific scenarios, they often depend on stable data distributions, predefined failure patterns, and sufficient labeled data \cite{trioxpert}. 
As a result, they suffer from limited generalizability and interpretability, making them difficult to apply reliably in real-world operations~\cite{opsagent}.

Recent advances in large language models (LLMs) have demonstrated remarkable capabilities in language understanding, knowledge utilization, and complex reasoning~\cite{InstructGPT, gpt4}.
With these capabilities, LLMs can process diverse operational information and perform diagnostic reasoning, providing new opportunities for improving failure diagnosis~\cite{flowxpert, flow-of-action, dbaiops, foundroot}.
In real-world operations, engineers diagnose failures through iterative evidence collection, observation analysis, and hypothesis refinement guided by operational experience \cite{gos}.

To support such iterative diagnosis, ReAct~\cite{react} enables LLMs to interleave reasoning and actions, forming a feedback loop between evidence acquisition and hypothesis refinement.
However, in traditional ReAct-style diagnosis, the evolving diagnostic trajectory is often accumulated as a linear sequence of thoughts, actions, and observations in the context window.
As the trajectory grows longer, such linear context becomes less reliable, making long-horizon diagnosis unstable \cite{lost-in-the-middle, llm-get-lost}.
GoS~\cite{gos} highlights typical long-horizon failures, including evidence fabrication, context drift, and failed backtracking, and mitigates them by organizing evidence and hypotheses into a structured belief state.
Nevertheless, reliable diagnosis still requires operational experience for guidance.

Although LLMs acquire broad parametric knowledge during training, such knowledge is often insufficient for real-world failure diagnosis, which relies heavily on system-specific operational experience \cite{RCACopilot}.
Existing methods usually incorporate such experience through retrieval, where a diagnosis query is used to retrieve relevant knowledge as auxiliary context.
Typical implementations include VectorRAG \cite{vector-rag}, which retrieves semantically similar document chunks, and graph-based RAG \cite{graph-rag,hippo-rag2 ,linear-rag}, which exploits structured relations for retrieval.
Regardless of the retrieval form, deciding what knowledge to expose to the context is critical: missing relevant knowledge provides limited guidance, while irrelevant knowledge causes context pollution \cite{pmlr-v202-shi23a}.
Moreover, as observations and hypotheses evolve during diagnosis, statically retrieved knowledge can be weakly aligned with the current diagnostic state.

As illustrated in Fig.~\ref{fig:teaser}, agentic-reasoning and knowledge-augmented methods address two complementary aspects of failure diagnosis.
Agentic-reasoning methods emphasize diagnostic state, while knowledge-augmented methods emphasize operational experience.
However, effective diagnosis requires the two to be tightly coupled: the current diagnostic state should activate relevant experience to guide reasoning.
For example, when current state contains evidence of sleeping database connections, it should activate experience about idle-slot occupation rather than generic overload. 
This motivates a dual-memory perspective.
Building such a framework faces challenges.
\textbf{(1) Memory Representation.} Modeling incident-specific states and cross-incident experience with different granularities.
\textbf{(2) Memory Coordination.} Bridging dynamic diagnostic states with relevant experience during reasoning.
\textbf{(3) Memory Consolidation.} Incorporating new experience while avoiding unreliable memory accumulation.

To address these challenges, we propose \name, a dual-memory framework for failure diagnosis.  
\name maintains two graph-structured memories: a short-term memory (STM) to capture the current diagnostic state and a long-term memory (LTM) to organize reusable operational experience.
During diagnosis, \name uses cross-memory resonance to activate LTM subgraphs relevant to the current STM, and then performs memory-conditioned diagnosis based on both the STM and the activated LTM.
After diagnosis, \name consolidates solved incidents into reusable experience, enabling the LTM to evolve while avoiding noise.
Our main contributions are summarized as follows:
\begin{itemize}
    \item To the best of our knowledge, \name is the first dual-memory framework that jointly models diagnostic state and operational experience for failure diagnosis.
    \item We introduce cross-memory resonance to activate state-relevant experience, enabling the diagnostic process conditioned on both STM and LTM.
    \item We design a long-term memory consolidation mechanism that distills solved incidents into reusable operational experience, allowing the LTM to continuously evolve.
    \item We evaluate \name on a real-world Huawei dataset, where it improves Match and Relevant by up to 46.88\% and 18.39\% over the strongest baseline, respectively.
    \item To support reproducibility, we make all code and prompts publicly available. \footnote{\url{https://github.com/gaorch85/OpsMem}} 
\end{itemize}


\section{Related Work}\label{sec:relate-works}

\noindent
\textbf{Agent-Based Failure Diagnosis.}
Recent studies have explored LLM-based agents for failure diagnosis.
\textsc{ReAct}~\cite{fse-react} explore ReAct-style agents for root cause analysis.
TrioXpert~\cite{trioxpert} employs multi-agent collaboration for anomaly detection, failure classification, and root cause localization, while FoundRoot\cite{foundroot} and R-Log\cite{R-Log} enhance LLM reasoning through structured reasoning and reinforcement learning, respectively.
GoS\cite{gos} further models diagnosis as an abductive reasoning task and maintains an explicit belief state to support failure diagnosis.
However, how to incorporate operational experience into diagnosis remains underexplored in these methods.

\noindent
\textbf{Knowledge-Augmented Failure Diagnosis.}
Another line of work augments diagnosis with operational knowledge.
RCACopilot\cite{RCACopilot} uses historical failure cases to support root cause analysis.
Flow-of-Action\cite{flow-of-action} and OScope\cite{OScope} further introduce SOP documents and historical cases to constrain diagnostic reasoning and improve interpretability.
DBAIOps\cite{dbaiops} represents database operational experience as a heterogeneous knowledge graph, while FlowXpert\cite{flowxpert} constructs hybrid knowledge bases for troubleshooting workflow generation.
However, the coordination between operational knowledge and the evolving diagnostic state remains insufficiently studied.
\section{Methodology}\label{sec:methodology}

\begin{figure*}[t]
    \centering
    \includegraphics[width=1\textwidth]{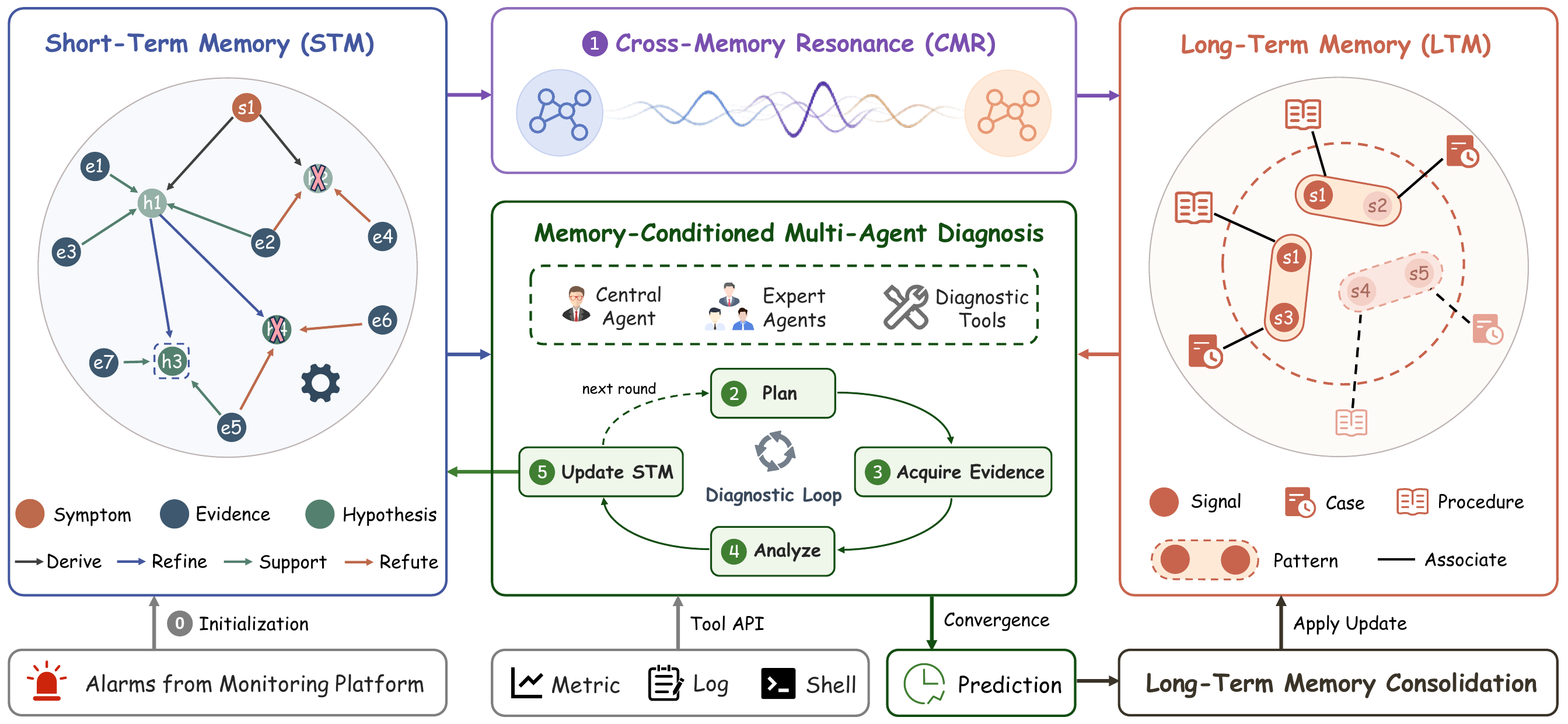}
    \vspace{-15pt}
    \caption{
        Overview of \name
    }
    \label{fig:overview}
    \vspace{-10pt}
\end{figure*}

We present \name, a novel dual-memory framework for failure diagnosis. 
As shown in Fig.~\ref{fig:overview}, given the alarms from the monitoring platform, \name first initializes a short-term memory (STM) to represent the evolving diagnostic state of the current incident. 
At each diagnostic round, the current STM triggers cross-memory resonance (CMR) over the long-term memory (LTM), which activates relevant operational experience for the current diagnostic state.
The current STM and the activated LTM then jointly condition a multi-agent diagnosis loop, where agents plan actions, acquire evidence, analyze observations, and update the STM. 
The updated STM further drives the next round of CMR, keeping the activated LTM aligned with the STM until convergence.
After the diagnosis is completed, a multi-agent consolidation module distills the reusable experience into LTM for future diagnosis.

The remainder of this section introduces the four core components of \name. 
Section~\ref{sec:methodology:dual-memory} describes the graph-structured representations of STM and LTM.
Section~\ref{sec:methodology:cmr} presents how CMR aligns the current diagnostic state with reusable operational experience. 
Section~\ref{sec:methodology:diagnosis} explains how dual memory conditions multi-agent diagnosis. 
Finally, Section~\ref{sec:methodology:consolidation} introduces the LTM consolidation mechanism.

\vspace{-5pt}
\subsection{Dual-Memory Architecture}\label{sec:methodology:dual-memory}

To support failure diagnosis, \name maintains two graph-structured memories: a short-term memory (STM) that captures the current diagnostic state, and a long-term memory (LTM) that organizes reusable operational experience.

\noindent\textbf{Short-Term Memory.} 
The STM is represented as a graph $\mathcal{M}_s=(\mathcal{V}_s,\mathcal{E}_s)$ that maintains the evolving diagnostic state of the current incident, following the belief-state abstraction in GoS~\cite{gos}.
The nodes in $\mathcal{V}_s$ include observed symptoms, acquired evidence, and candidate hypotheses. 
The edges in $\mathcal{E}_s$ encode diagnostic relations among them, such as derive, refine, support, and refute. 
During diagnosis, newly acquired evidence and intermediate analyses are continuously written into the STM, making it an explicit memory of what has been observed, inferred, and ruled out so far.

\noindent\textbf{Long-Term Memory.}
The LTM is represented as another graph $\mathcal{M}_l=(\mathcal{V}_l,\mathcal{E}_l)$ that organizes cross-incident operational experience.
The node set $\mathcal{V}_l$ contains three types of nodes: patterns, cases, and procedures.
A pattern is the core node in the LTM, representing a typical failure mode as a structured association between diagnostic signals, where each signal is a normalized cue that describes a system condition, such as high memory usage.
A case records a historical incident related to a pattern, while a procedure records diagnostic steps for that pattern.
The weighted edge set $\mathcal{E}_l$ associates patterns with cases and procedures, enabling matched patterns to activate both historical examples and actionable diagnostic guidance.
Unlike the STM, the LTM accumulates cross-incident experience and can be repeatedly activated during diagnosis.

\subsection{Cross-Memory Resonance}\label{sec:methodology:cmr}

\begin{figure*}[t]
    \centering
    \includegraphics[width=1\textwidth]{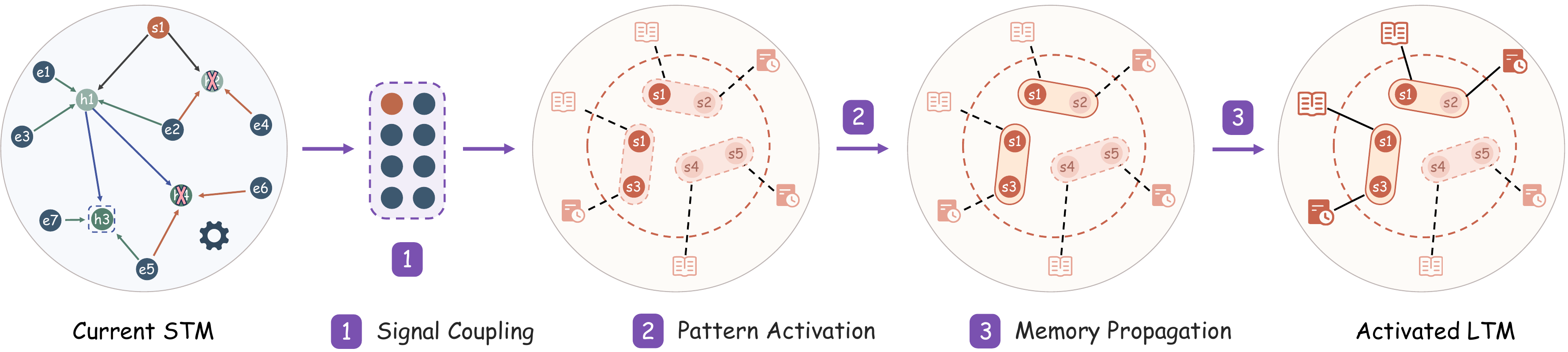}
    \vspace{-14pt}
    \caption{
        Cross-Memory Resonance
    }
    \label{fig:cmr}
    \vspace{-11pt}
\end{figure*}

As shown in Fig.~\ref{fig:cmr}, cross-memory resonance (CMR) aligns the diagnostic state with operational experience. 
It consists of three steps: signal coupling, pattern activation, and memory propagation, which together produce an activated LTM subgraph for the subsequent diagnosis loop.

\noindent\textbf{Signal Coupling.} 
CMR first extracts symptom and evidence nodes from the STM and normalizes them into query signals through an LLM.
These query signals are then matched with the signals in LTM patterns.
Each LTM signal receives a coupling score based on its maximum semantic similarity to the query signals, and scores below a predefined threshold are set to zero, while the remaining signals are treated as active.
This step grounds LTM activation in the concrete observations already recorded in the current STM.

\noindent\textbf{Pattern Activation.} 
After signal coupling, CMR lifts signal-level scores to pattern-level activation. 
For each pattern, CMR computes its activation score from two factors: the average score of active signals and the ratio of active signals within the pattern.
The two factors respectively capture the alignment strength of activated signals and the coverage of the pattern by current observations.
Only patterns whose activation scores exceed a predefined threshold are retained for subsequent propagation.
In this way, LTM activation focuses on recurring diagnostic associations rather than isolated observations.

\noindent\textbf{Memory Propagation.} 
Starting from the activated patterns, CMR propagates activation to associated nodes according to the edge weights. 
The propagation scores are used to select the most relevant cases and procedures. 
Together, activated patterns suggest diagnostic directions, cases provide historical references, and procedures guide subsequent diagnosis.

Whenever the STM is updated during diagnosis, CMR is applied again to refresh the activated LTM subgraph according to the latest diagnostic state.

\subsection{Memory-Conditioned Multi-Agent Diagnosis}\label{sec:methodology:diagnosis}

Given the current STM and the activated LTM subgraph, \name performs a memory-conditioned diagnosis loop.
Both memories are serialized into each agent's prompt to jointly condition agent actions.

Each diagnostic round consists of four steps.
First, in \emph{Plan}, the CentralAgent plans one or more diagnostic tasks and assigns them to selected ExpertAgents.
Second, in \emph{Acquire Evidence}, the ExpertAgents iteratively invoke diagnostic tools, such as a log retrieval, to collect observations.
Third, in \emph{Analyze}, the ExpertAgents interpret the collected observations and return concise diagnostic reports to the CentralAgent.
Finally, in \emph{Update STM}, the CentralAgent aggregates these reports and updates the STM with new evidence, hypotheses, and diagnostic relations.

After each STM update, \name checks for convergence following GoS~\cite{gos}.
If a sufficiently supported hypothesis is identified, \name outputs the prediction.
Otherwise, the updated STM triggers another round of CMR to refresh the activated LTM subgraph for the next diagnostic round.

\vspace{-5pt}
\subsection{Long-Term Memory Consolidation}\label{sec:methodology:consolidation}

\begin{figure}[htbp]
    \vspace{-10pt}
    \centering
    \includegraphics[width=\columnwidth]{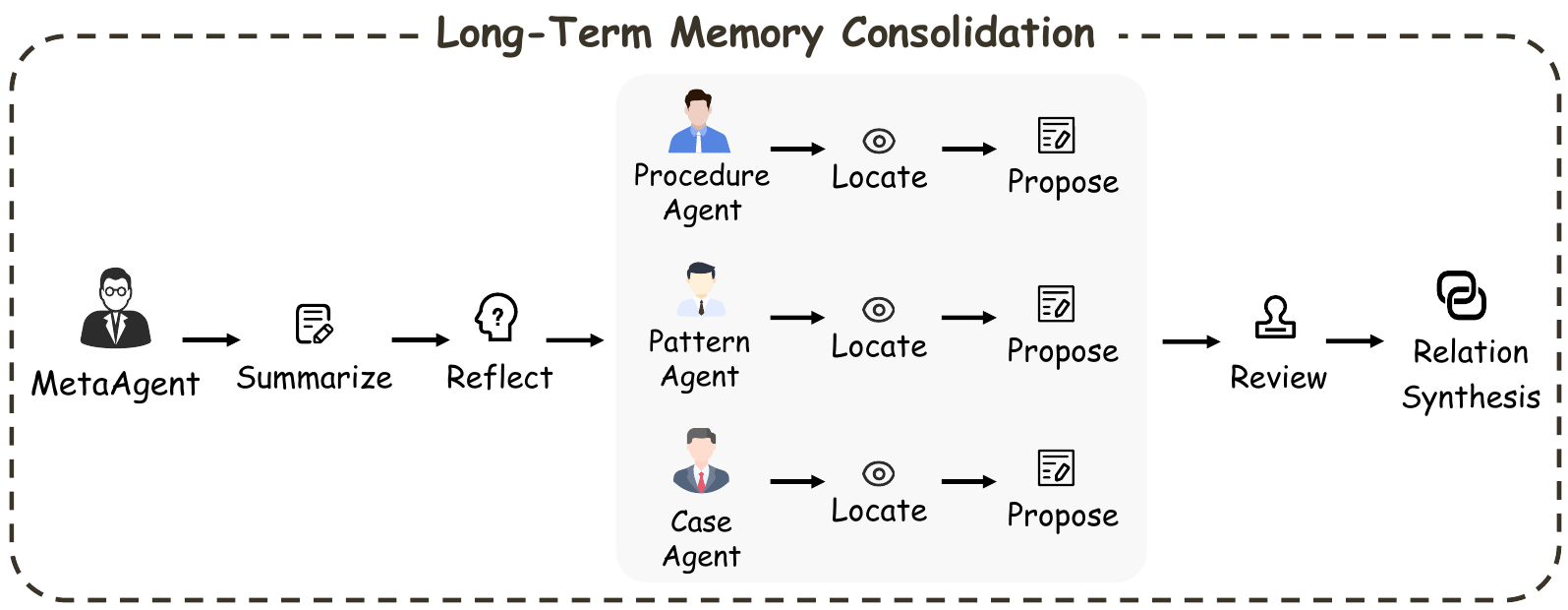}
    \vspace{-18pt}
    \caption{
        Long-Term Memory Consolidation
    }
    \label{fig:consolidation}
    \vspace{-5pt}
\end{figure}

After a failure is resolved, \name invokes another multi-agent consolidation module to update the LTM.
As shown in Fig.~\ref{fig:consolidation}, consolidation is coordinated by MetaAgent, which first summarizes the diagnostic trace and reflects on it with the final STM and the activated LTM subgraph.
Based on this reflection, the MetaAgent identifies which memories need updates and invokes the corresponding memory agents.

For each invoked memory agent, consolidation first locates related memory nodes in the current LTM, using the activated LTM subgraph as the context.
The agent then proposes node-level operations based on the diagnosis summary, the final STM, and the located memory nodes.
Each proposal consists of \textsc{Create} and \textsc{Delete} operations: \textsc{Create} adds new nodes, while \textsc{Delete} removes obsolete nodes.
When an existing memory node needs revision, the agent deletes the old version and creates a revised one.

The MetaAgent reviews all proposals to filter out unqualified updates.
Finally, relation synthesis integrates the validated memory nodes into the LTM through new edges, making them reusable in future incidents.

\section{Experiments}\label{sec:experiment}
In this section, we aim to answer the following research questions (RQs):\\
\textbf{RQ1:} How effective is \name for failure diagnosis?\\
\textbf{RQ2:} Does each component contribute to \name?\\
\textbf{RQ3:} Can \name improve over time?

\subsection{Experimental Setup}

\subsubsection{Datasets}
To comprehensively evaluate \name under realistic failure diagnosis scenarios, we construct a dataset from Huawei's production microservice systems.
The dataset contains 120 real-world failure incidents, covering a broad spectrum of fault types across application services, resource management, databases, and infrastructures.
For each incident, we collect the initial failure alarms and multi-source diagnostic observations obtained during incident handling, such as time-series metrics, system logs, and shell-level snapshots.
The label is confirmed by on-call engineers through post-incident analysis and used as the ground truth for evaluation.

\subsubsection{Evaluation}
We adopt an LLM-as-a-Judge protocol with Qwen3.5-27B to evaluate diagnostic effectiveness.
Following GoS~\cite{gos}, the judge compares each predicted root cause with the ground truth on a three-point scale: 2 for exact match, 1 for relevant, and 0 otherwise.
We report \textit{Match} as the proportion of predictions scored 2, and \textit{Relevant} as the proportion of predictions scored at least 1.
To improve reliability, we apply self-consistency~\cite{sc-cot} by obtaining at least five independent judgments and aggregating them via majority voting, and further validate the results by expert sampling.

\subsubsection{Baselines}
We compare \name with two groups of representative baselines.
For agentic-reasoning methods, we include ReAct~\cite{react}, which interleaves reasoning and actions, and GoS~\cite{gos}, which maintains a belief state for long horizon diagnosis.
For knowledge-augmented methods, we equip GoS with three representative RAG strategies.
VectorRAG \cite{vector-rag} retrieves semantically similar chunks from the corpus.
GraphRAG \cite{graph-rag} builds an entity graph and uses community summaries for retrieval.
LinearRAG \cite{linear-rag} constructs a relation-free graph and retrieves passages through entity activation.

\subsubsection{Long-Term Memory Construction}
Before evaluation, \name is initialized with an LTM built from interviews, questionnaires, and operational documents.
We use GPT-5.4 to assist in extracting failure patterns, cases, and procedures from these sources.
The LTM sources were finalized before evaluation, with all 120 evaluation incidents excluded.
For fair comparison, all knowledge-augmented baselines use the same knowledge sources but organize them in their own retrieval formats.

\subsubsection{Implementation Details}
For signal coupling, we encode signals with BGE-M3 and compute cosine similarity between embeddings.
Both the signal and pattern activation thresholds are set to 0.6.
The two factors in pattern activation are equally weighted.
At each CMR step, we retain the top-3 patterns, cases, and procedures.
The diagnosis loop runs up to 3 rounds and returns the best-supported hypothesis if not converged.

\subsection{RQ1: Overall Performance}

\begin{table}[hbpt]
    \vspace{-10pt}
    \centering
    \caption{Overall Performance (\%)}
    \label{tab:overall}
    \renewcommand{\arraystretch}{1.12}
    \setlength{\tabcolsep}{3pt}
    \begin{tabular*}{\columnwidth}{@{}l@{\hspace{1.5em}}l@{\hspace{7em}}c@{\extracolsep{\fill}}c@{}}
        \toprule
        \textbf{Seed LLM} & \textbf{Method} & \textbf{Match} & \textbf{Relevant} \\
        \midrule
        \multirow{8}{*}{Qwen3.5-27B}
            & \multicolumn{3}{l}{\emph{Agentic-Reasoning Methods}} \\
            & \quad ReAct & 20.83 & 49.17 \\
            & \quad GoS & 30.83 & 55.83 \\
            & \multicolumn{3}{l}{\emph{Knowledge-Augmented Methods}} \\
            & \quad GoS + VectorRAG & 48.33 & 61.67 \\
            & \quad GoS + GraphRAG & 53.33 & 71.67 \\
            & \quad GoS + LinearRAG & 53.33 & 72.50 \\
            & \textbf{\name} & \textbf{78.33} & \textbf{85.83} \\
        \cmidrule(lr){1-4}
        \multirow{8}{*}{Gemma-4-31B}
            & \multicolumn{3}{l}{\emph{Agentic-Reasoning Methods}} \\
            & \quad ReAct & 22.50 & 58.33 \\
            & \quad GoS & 29.17 & 51.67 \\
            & \multicolumn{3}{l}{\emph{Knowledge-Augmented Methods}} \\
            & \quad GoS + VectorRAG & 50.00 & 73.33 \\
            & \quad GoS + GraphRAG & 56.67 & 79.17 \\
            & \quad GoS + LinearRAG & 48.33 & 72.50 \\
            & \textbf{\name} & \textbf{63.33} & \textbf{82.50} \\
        \cmidrule(lr){1-4}
        \multirow{8}{*}{GLM-4-32B}
            & \multicolumn{3}{l}{\emph{Agentic-Reasoning Methods}} \\
            & \quad ReAct & 17.50 & 42.50 \\
            & \quad GoS & 21.67 & 59.17 \\
            & \multicolumn{3}{l}{\emph{Knowledge-Augmented Methods}} \\
            & \quad GoS + VectorRAG & 34.17 & 66.67 \\
            & \quad GoS + GraphRAG & 33.33 & 65.83 \\
            & \quad GoS + LinearRAG & 40.00 & 69.17 \\
            & \textbf{\name} & \textbf{53.33} & \textbf{77.50} \\
        \bottomrule
    \end{tabular*}
\end{table}

Table~\ref{tab:overall} reports the overall diagnosis performance.
Across all three seed LLMs, \name consistently achieves the best results on both Match and Relevant, showing that its benefit is not tied to a specific backbone.
Compared with the strongest baseline under each seed LLM, \name improves Match by 6.66--25.00 points and Relevant by 3.33--13.33 points.

Agentic-Reasoning methods, such as ReAct \cite{react} and GoS \cite{gos}, perform worse because they mainly organize the current reasoning process but lack operational experience.
Knowledge-Augmented variants improve over GoS, confirming the value of external experience.
However, they still lag behind \name, since retrieved knowledge is not explicitly aligned with the evolving diagnostic state.
By maintaining the current diagnostic state in STM and activating state-relevant experience from LTM through CMR, \name better guides evidence acquisition and hypothesis evaluation, leading to more accurate root-cause identification.

\subsection{RQ2: Ablation Study}

\begin{table}[!hbpt]
    \vspace{-10pt}
    \centering
    \caption{Ablation Study (\%)}
    \label{tab:ablation}
    \renewcommand{\arraystretch}{1.12}
    \setlength{\tabcolsep}{3pt}
    \begin{tabular*}{\columnwidth}{@{}l@{\hspace{3.5em}}l@{\hspace{3.5em}}c@{\extracolsep{\fill}}c@{}}
        \toprule
        \textbf{Seed LLM} & \textbf{Method} & \textbf{Match} & \textbf{Relevant} \\
        \midrule
        \multirow{5}{*}{Qwen3.5-27B}
            & \textbf{\name} & \textbf{78.33} & \textbf{85.83} \\
            & w/o STM & 45.00 & 63.33 \\
            & w/o LTM & 30.83 & 55.83 \\
            & w/o CMR & 56.67 & 68.33 \\
            & w/o LTM Consolidation & 70.83 & 80.83 \\
        \bottomrule
    \end{tabular*}
\end{table}

Table~\ref{tab:ablation} validates the contribution of each component.
Removing LTM causes the largest degradation, confirming the importance of operational experience.
Removing STM also substantially hurts performance, indicating that explicit state tracking is necessary to maintain accumulated evidence and hypotheses during long-horizon diagnosis.
The drop without CMR further shows the importance of state-aware LTM activation, which aligns operational experience with the evolving diagnostic state.
Removing LTM consolidation yields a smaller but consistent decline, suggesting that LTM evolution brings additional gains.
Overall, both memories, their coordination, and consolidation are necessary for \name.

\subsection{RQ3: Self-Evolution via Long-Term Memory Consolidation}

\begin{table}[hbpt]
    \vspace{-10pt}
    \centering
    \caption{Self-Evolution Across Incident Windows}
    \label{tab:self_evolution}
    \renewcommand{\arraystretch}{1.12}
    \newcommand{\crange}[2]{%
        \makebox[5.0em][c]{%
            \makebox[1.4em][c]{#1}%
            \hfil--\hfil%
            \makebox[1.8em][c]{#2}%
        }%
    }
    \begin{tabular*}{\columnwidth}{@{\extracolsep{\fill}} l c >{\hspace{2em}}c >{\hspace{-1.50em}}c @{}}
        \toprule
        \multirow{2}{*}{\textbf{Seed LLM}}
        & \multirow{2}{*}{\textbf{Incident Range}}
        & \multicolumn{2}{c}{\textbf{Gains vs. w/o LTM Con.}} \\
        \cmidrule(lr){3-4}
        &
        & \textbf{Match} & \textbf{Relevant} \\
        \midrule
        \multirow{4}{*}{Qwen3.5-27B}
            & \crange{1}{30}    & +0 & +1 \\
            & \crange{31}{60}   & +3 & +1 \\
            & \crange{61}{90}   & +1 & +3 \\
            & \crange{91}{120}  & +5 & +1 \\
        \bottomrule
    \end{tabular*}
\end{table}
Table~\ref{tab:self_evolution} further examines whether LTM consolidation enables \name to improve over time.
We process the 120 incidents sequentially and split them into four consecutive windows.
After each diagnosed incident, \name consolidates validated experience into the LTM, while the w/o LTM Consolidation variant keeps the initial LTM fixed.
Compared with this variant, \name correctly diagnoses additional incidents under both Match and Relevant metrics in every window, with the largest exact-match gain appearing in the last window.
This suggests that consolidated experience can be reused by subsequent diagnoses, allowing the LTM to evolve over time.

\subsection{Case Study}
\begin{figure}[htbp]
    \vspace{-5pt}
    \centering
    \includegraphics[width=\columnwidth]{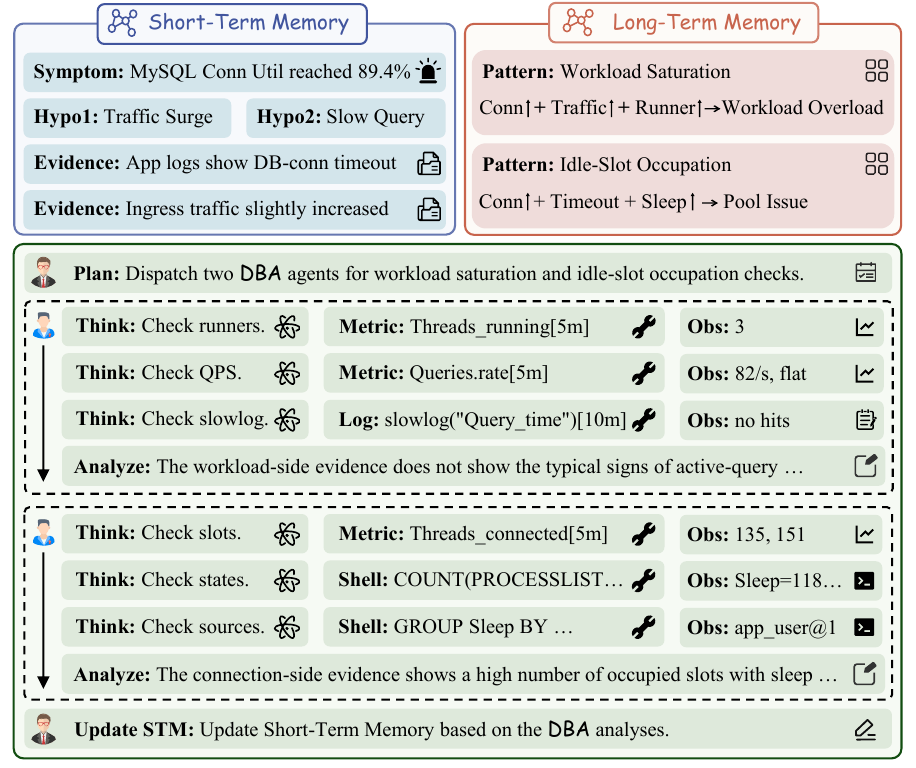}
    \vspace{-18pt}
    \caption{A diagnosis round of \name on a real-world incident}
    \label{fig:case}
    \vspace{-10pt}
\end{figure}

Fig.~\ref{fig:case} shows one diagnosis round of \name on an anonymized real-world Huawei incident.
Starting from the current STM, \name activates two relevant LTM patterns, workload saturation and idle-slot occupation, and dispatches two DBA agents for targeted checks.
The workload-side evidence weakens the overload/slow-query hypotheses, while the connection-side evidence supports idle-slot occupation caused by connection-pool issues.
\name then updates the STM with the new evidence and analysis.

\section{Conclusion}\label{sec:conclusion}

This paper presents \name, a dual-memory framework for failure diagnosis that couples the short-term diagnostic memory with long-term operational memory. 
\name uses Cross-Memory Resonance to align the current diagnostic state with relevant operational experience, and conditions the diagnosis on both memories.
After a failure is resolved, \name distills reusable experience into LTM, enabling \name to evolve over time.
Experiments on a real-world Huawei microservice dataset show that \name consistently outperforms representative baselines, and further studies validate the effectiveness of dual-memory coordination and LTM consolidation.

\bibliographystyle{IEEEtran}
\bibliography{IEEEabrv,bibliography}

\end{document}